\pdfoutput=1

\documentclass[letterpaper, 10pt, conference]{ieeeconf}   %

\IEEEoverridecommandlockouts                              %

\let\labelindent\relax                                    %

\usepackage{graphicx}     %
\usepackage{amsmath}      %
\usepackage{amssymb}      %
\usepackage{amsfonts}     %
\usepackage{mathtools}    %
\usepackage{enumitem}     %
\usepackage{siunitx}      %
\usepackage{tensor}       %
\usepackage{url}          %
\usepackage{xcolor}       %
\usepackage{xspace}       %
\usepackage[T1]{fontenc}  %
\usepackage{hyperref}     %
\usepackage[all]{hypcap}  %

\hypersetup{hidelinks, bookmarksdepth=3, bookmarksopen, bookmarksnumbered, colorlinks=false, linkcolor={red!50!black}, citecolor={blue!50!black}, urlcolor={blue!80!black}}

\makeatletter
\def\subsubsection{\@startsection{subsubsection}{3}{\parindent}{0ex plus 0.1ex minus 0.1ex}{0ex}{\normalfont\normalsize\bfseries}}
\makeatother

\makeatletter
\newcommand*\bigcdot{\mathpalette\bigcdot@{.5}}
\newcommand*\bigcdot@[2]{\mathbin{\vcenter{\hbox{\scalebox{#2}{$\m@th#1\bullet$}}}}}
\makeatother

\graphicspath{{images/}}
\DeclareGraphicsExtensions{.pdf,.png,.jpg,.jpeg}

\DeclareMathOperator{\sign}{sign}
\DeclareMathOperator{\asin}{asin}
\DeclareMathOperator{\acos}{acos}
\DeclareMathOperator{\wrap}{wrap}

\DeclareMathOperator{\atantwo}{atan2}
\providecommand{\abs}[1]{\left\lvert#1\right\rvert}

\newcommand{\vect}[1]{\mathbf{#1}} %
\newcommand{\vecs}[2]{\tensor*{\vect{#1}}{_{#2}}} %
\newcommand{\compbs}[3]{\tensor*[^{#1}]{#2}{_{#3}}} %
\newcommand{\rots}[2]{\tensor*{#1}{_{#2}}} %
\newcommand{\rotb}[3]{\tensor*[^{#1}_{#2}]{#3}{}} %
\newcommand{\yawof}[1]{\Psi\bigl(#1\bigr)} %
\newcommand{\fr}[1]{\{#1\}\xspace} %
\newcommand{\inv}{^{-1}} %
\newcommand{\conj}{^{*}} %
\newcommand{\trans}{^T\!} %
\newcommand{\E}{\mathbb{E}} %
\newcommand{\Et}{\mathbb{\tilde{E}}} %
\newcommand{\F}{\mathbb{F}} %
\newcommand{\I}{\mathbb{I}} %
\newcommand{\Q}{\mathbb{Q}} %
\newcommand{\R}{\mathbb{R}} %
\newcommand{\T}{\mathbb{T}} %
\newcommand{\half}{\tfrac{1}{2}} %
\newcommand{\hpi}{\tfrac{\pi}{2}} %
\newcommand{\defeq}{\equiv} %
\newcommand{\degreem}{^{\circ}} %
\newcommand{\sub}[2]{#1_{\scriptscriptstyle\mspace{-2mu}#2}}
\newcommand{\psiE}{\sub{\psi}{E}}
\newcommand{\psibE}{\sub{\bar{\psi}}{E}}
\newcommand{\psiEt}{\sub{\psi}{\widetilde{E}}}
\newcommand{\psibEt}{\sub{\bar{\psi}}{\widetilde{E}}}
\newcommand{\thE}{\sub{\theta}{E}}
\newcommand{\thbE}{\sub{\bar{\theta}}{E}}
\newcommand{\thEt}{\sub{\theta}{\widetilde{E}}}
\newcommand{\thbEt}{\sub{\bar{\theta}}{\widetilde{E}}}
\newcommand{\phiE}{\sub{\phi}{E}}
\newcommand{\phibE}{\sub{\bar{\phi}}{E}}
\newcommand{\phiEt}{\sub{\phi}{\widetilde{E}}}
\newcommand{\phibEt}{\sub{\bar{\phi}}{\widetilde{E}}}

\newcommand{\seclabel}[1]{\label{sec:#1}}

\newcommand{\figlabel}[1]{\label{fig:#1}}
\newcommand{\tablabel}[1]{\label{tab:#1}}
\newcommand{\eqnlabel}[1]{\label{eqn:#1}}
\newcommand{\enumlabel}[1]{\label{enum:#1}}
\newcommand{\secref}[1]{Section~\ref{sec:#1}\xspace}

\newcommand{\figref}[1]{Fig.~\ref{fig:#1}\xspace}
\newcommand{\tabref}[1]{Table~\ref{tab:#1}\xspace}
\newcommand{\eqnref}[1]{(\ref{eqn:#1})\xspace}
\newcommand{\enumref}[1]{\ref{enum:#1}\xspace}

\newcommand{\eqnrefs}[2]{(\ref{eqn:#1}--\ref{eqn:#2})\xspace}

\newcommand{\cpp}{C\texttt{\nolinebreak\hspace{-.05em}+\nolinebreak\hspace{-.05em}+}\xspace}
\newcommand{\degree}{$\degreem$\xspace}

\title{\LARGE \bf Fused Angles and the Deficiencies of Euler Angles}

\author{Philipp Allgeuer and Sven Behnke%
\thanks{All authors are with the Autonomous Intelligent Systems (AIS) Group, Computer Science Institute VI,
        University of Bonn, Germany. Email: {\tt\small pallgeuer@ais.uni-bonn.de}. This work was partially
        funded by grant BE 2556/13 of the German Research Foundation (DFG).}}

\usepackage{eso-pic}

\AtBeginDocument{\AddToShipoutPictureFG*{\AtTextUpperLeft{\put(0,\LenToUnit{9pt}){\parbox{\textwidth}{\centering\bfseries
International Conference on Intelligent Robots and Systems (IROS), Madrid, Spain, 2018
}}}}}
\begin{document}
\begin{allowdisplaybreaks}

\bstctlcite{IEEEexample:BSTcontrol}

\maketitle
\thispagestyle{empty}
\pagestyle{empty}

\begin{abstract}
Just like the well-established Euler angles representation, fused angles are a 
convenient parameterisation for rotations in three-dimensional Euclidean space. 
They were developed in the context of balancing bodies, most specifically 
walking bipedal robots, but have since found wider application due to their 
useful properties. A comparative analysis between fused angles and Euler angles 
is presented in this paper, delineating the specific differences between the two 
representations that make fused angles more suitable for representing 
orientations in balance-related scenarios. Aspects of comparison include the 
locations of the singularities, the associated parameter sensitivities, the 
level of mutual independence of the parameters, and the axisymmetry of the 
parameters.
\end{abstract}

\section{Introduction}
\seclabel{introduction}

The fused angles rotation parameterisation was recently introduced in 
\cite{Allgeuer2015a}. While it arose from the analysis and control of balancing 
bodies in 3D and has been used extensively as such \cite{Allgeuer2016a}, it has 
also since been used for various other purposes, including for example attitude 
estimation \cite{Allgeuer2014} and the modelling of foot orientations and ground 
contacts \cite{Farazi2016}. Libraries have also been released in \cpp 
\cite{RotConvLibGithub} and Matlab \cite{MatOctRotLibGithub} that implement a 
wide variety of conversions and operations involving fused angles and all of the 
classic ways of representing rotations.

Fused angles aim to provide a robust and geometrically intuitive way of 
quantifying the amount of rotation that a body has within each of the three 
major planes, i.e.\ the $\vect{xy}$, $\vect{yz}$ and $\vect{xz}$ planes, as 
illustrated on the left in \figref{fused_euler_teaser}. This can conceptually be 
thought of as requiring a notion of how to concurrently quantify the `amount of 
rotation' a body has about the three principal axes. Furthermore, it is an aim 
that the three quantified planar rotation values describe the state of balance 
in an intuitive, problem-relevant and symmetrical way, in particular with 
respect to the lateral, sagittal and transverse planes. Clearly, quaternions do 
not satisfy these stated aims as no three quaternion components directly 
quantify planar angles of rotation. Note that due to the balance-inspired nature 
of the task, the only required axiom is that there is some clear notion of `up'. 
This is generally the opposite direction to gravity, or along a particular 
surface normal, and without loss of generality is chosen to be represented by 
the global z-axis.

Euler angles, illustrated on the right in \figref{fused_euler_teaser}, may at 
first seem to satisfy these requirements, being a commonly accepted catch-all 
solution, but this is not entirely so. They can often enough be the correct 
choice for a task, such as for example for the modelling of gimbals, or a 
colocated series of joints, but too often they are chosen simply because there 
does not seem to be a reasonable alternative. This paper critically assesses 
Euler angles in direct comparison to fused angles, to elucidate the differences 
that make fused angles the more suitable candidate for quantifying the 
orientation of a balancing body. This comparative analysis, in addition to the 
presentation of some noteworthy properties of fused angles, is the main 
contribution of this paper.

\begin{figure}[!t]
\parbox{\linewidth}{\vspace*{-0.8ex}\centering\includegraphics[width=0.95\linewidth]{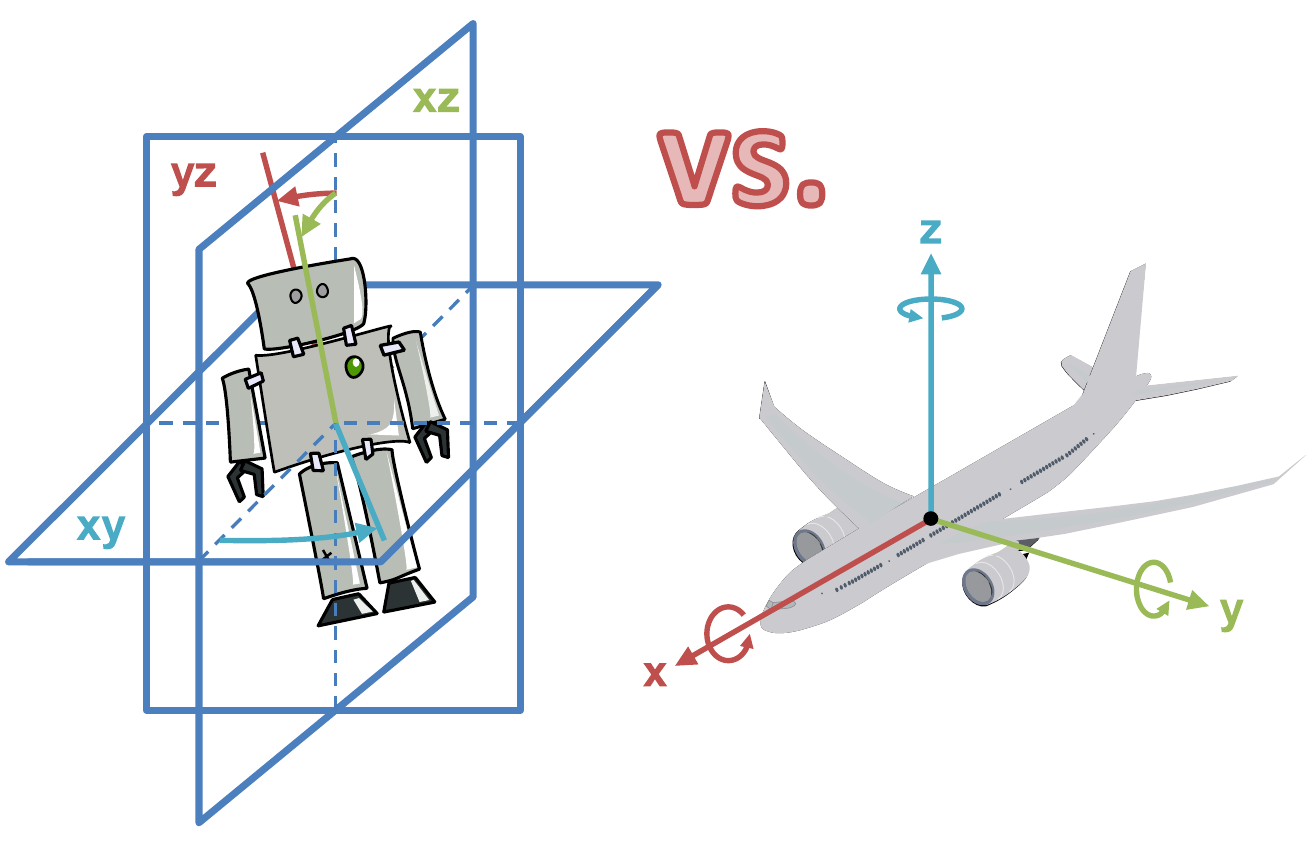}}
\vspace{-2.2ex}
\caption{Comparison of the fundamental principles of fused angles (left) and 
Euler angles (right). Fused angles quantify the amount of rotation within the 
three major planes, while Euler angles define three sequential axis rotations, 
e.g. around the z-axis, y-axis and then the x-axis.}
\figlabel{fused_euler_teaser}
\vspace{-2ex}
\end{figure}
\section{Related Work}
\seclabel{related_work}

Based on the aims that were outlined in \secref{introduction}, one can quickly 
see why existing rotation representations such as quaternions, rotation 
matrices, axis-angle pairs \cite{Palais2009}, rotation vectors 
\cite{Tomazic2011}, and vectorial parameterisations \cite{Bauchau2003} 
\cite{Trainelli2004}, are not appropriate for the task. For example, neither 
rotation matrices nor quaternions clearly identify components of rotation within 
the three major planes. A thorough review of all of these representations, and 
exactly why they are not suitable, can be found in \cite{Allgeuer2015a}. The 
only remaining possibly suitable classical rotation representation is Euler 
angles, which at least at first glance seems to satisfy the specified aims. It 
is briefly explained in \cite{Allgeuer2015a} why also this representation does 
not suffice for the required application, but providing a complete 
substantiation of this claim is the central topic of this paper.

\section{Review of Euler Angles}
\seclabel{review_euler}

Euler angles express a rotation as a sequence of three elemental rotations about 
a predefined set of coordinate axes, in a predefined order. The elemental 
rotations are either by convention extrinsic about the fixed global x, y and 
z-axes, or intrinsic about the local x, y and z-axes of the coordinate frame 
being rotated. For each of these two types, the order of axis rotations leads to 
six possible conventions where each axis is used only once, referred to as 
Tait-Bryan angles, and a further six possible conventions where the first and 
third axes of rotation are the same, referred to as proper Euler angles. All 
possible axis conventions are summarised in \tabref{euler_conventions}.

\begin{table}[!t]
\renewcommand{\arraystretch}{1.3}
\setlength\abovecaptionskip{0pt}
\caption{Complete list of Euler angles axis conventions}
\tablabel{euler_conventions}
\centering
\begin{tabular}{r l}
\hline
\textbf{Type} & \textbf{Order of axis rotations}\\
\hline
Proper Euler angles & XYX, XZX, YXY, YZY, ZXZ, ZYZ \\
Tait-Bryan angles & XYZ, XZY, YXZ, YZX, ZXY, ZYX \\
\hline
\end{tabular}
\vspace{-2ex}
\end{table}

It is easy to see that all extrinsic Euler angles conventions are completely 
equivalent to the corresponding intrinsic Euler angles conventions, just with 
the order of rotations reversed. As such, for comparison with fused angles, 
without loss of generality, intrinsic Euler angles are chosen. It is also 
desired for the three elemental rotations to be about three different axes, 
i.e.\ Tait-Bryan angles, so that the amount of rotation within each of the three 
major planes can be quantified. Furthermore, fused angles have an initial yaw 
rotation component about the z-axis, as detailed later in \secref{fused_angles}, 
so to facilitate a sensible comparison, only the intrinsic ZYX and ZXY Euler 
angles conventions remain as viable candidates. For completeness, both of these 
Euler angles conventions are presented briefly in the following sections, but 
unless explicitly otherwise stated, all further references to `Euler angles' 
will be referring to intrinsic ZYX Euler angles. All arguments and properties 
that apply to the ZYX representation can also equivalently be reformulated to 
apply to the ZXY representation, so the choice is arbitrary.

\subsection{Intrinsic ZYX Euler Angles}
\seclabel{euler_zyx}

Let \fr{G} denote a global reference frame, and \fr{B} be the frame of which the 
orientation is being expressed. The intrinsic ZYX Euler angles representation 
consists of the following three sequential rotations: first a rotation by the 
\emph{Euler yaw} $\psiE$ about the z-axis, then by the \emph{Euler pitch} $\thE$ 
about the new y-axis, and then by the \emph{Euler roll} $\phiE$ about the newest 
x-axis, as illustrated in \figref{euler_parameters}. The complete Euler angles 
rotation from \fr{G} to \fr{B} is then denoted by
\begin{equation}
\rotb{G}{B}{E} = (\psiE, \thE, \phiE) \in (-\pi,\pi] {\times} [-\hpi,\hpi] {\times} (-\pi,\pi] \equiv \E. \eqnlabel{eulerzyxdefn}
\end{equation}
The representation is unique, except at \emph{gimbal lock}, which is when $\thE 
= \pm\hpi$. The rotation matrix $R$ corresponding to the Euler angles rotation 
$E = (\psiE, \thE, \phiE)$ is given by
\begin{align}
R &= R_z(\psiE) R_y(\thE) R_x(\phiE) \eqnlabel{eulerzyxtorotmat} \\
&=
\begin{bmatrix}
c_{\psiE} c_{\thE} & c_{\psiE} s_{\thE} s_{\phiE} {-} s_{\psiE} c_{\phiE} & c_{\psiE} s_{\thE} c_{\phiE} {+} s_{\psiE} s_{\phiE} \\
s_{\psiE} c_{\thE} & s_{\psiE} s_{\thE} s_{\phiE} {+} c_{\psiE} c_{\phiE} & s_{\psiE} s_{\thE} c_{\phiE} {-} c_{\psiE} s_{\phiE} \\
-s_{\thE} & c_{\thE} s_{\phiE} & c_{\thE} c_{\phiE}
\end{bmatrix}\mspace{-8mu},\mspace{-2mu} \notag
\end{align}
where $s_{\ast} \equiv \sin(\ast)$ and $c_{\ast} \equiv \cos(\ast)$. The 
conversion from Euler angles to quaternion form $q = (w,x,y,z)$ is given by
\begin{equation}
\begin{split}
\mspace{-6mu}q = (&c_{\phibE} c_{\thbE} c_{\psibE} \!+ s_{\phibE} s_{\thbE} s_{\psibE}, s_{\phibE} c_{\thbE} c_{\psibE} \!- c_{\phibE} s_{\thbE} s_{\psibE}, \\
&c_{\phibE} s_{\thbE} c_{\psibE} \!+ s_{\phibE} c_{\thbE} s_{\psibE}, c_{\phibE} c_{\thbE} s_{\psibE} \!- s_{\phibE} s_{\thbE} c_{\psibE}),\mspace{-5mu}
\end{split}
\eqnlabel{eulerzyxtoquat}
\end{equation}
where for example $s_{\thbE} = \sin{\thbE} = \sin(\half\thE)$.

\begin{figure}[!t]
\parbox{\linewidth}{\centering\includegraphics[width=1.00\linewidth]{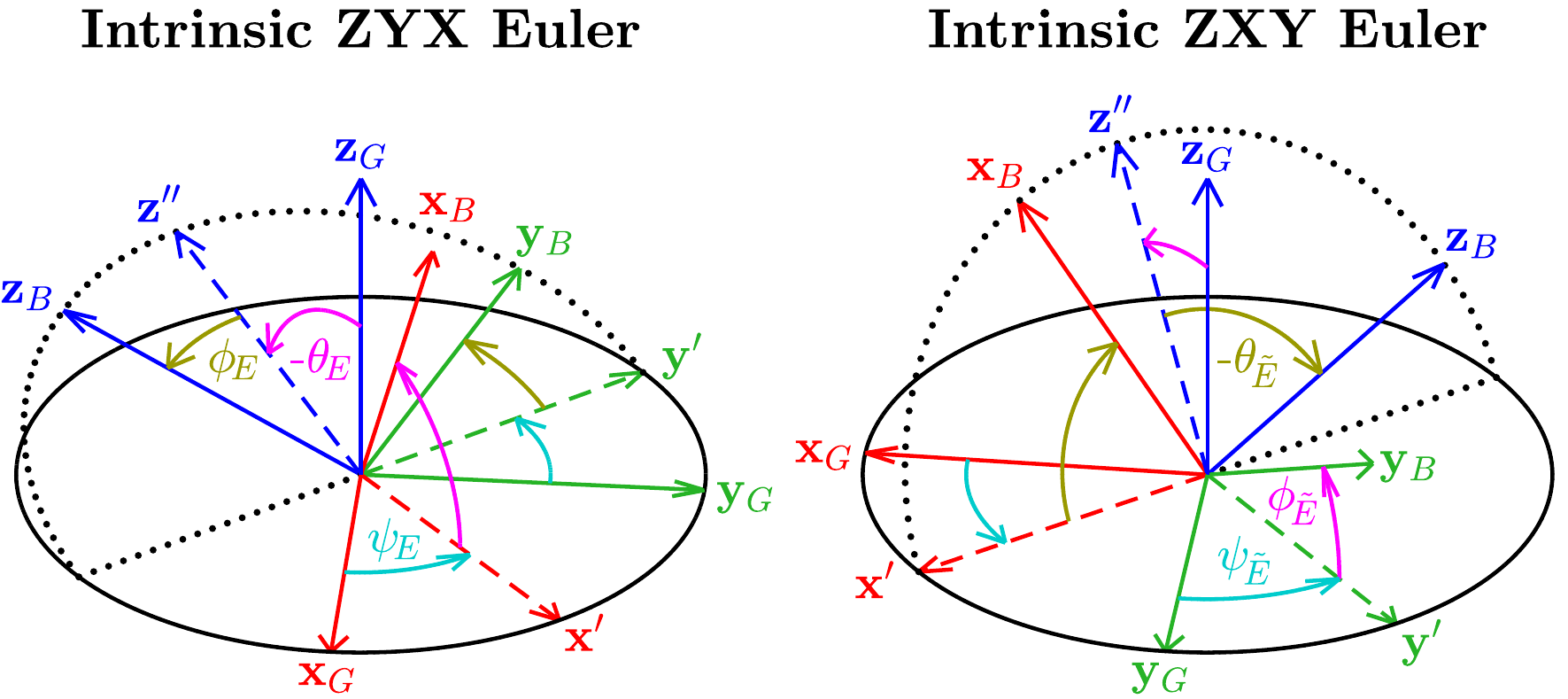}}
\caption{Diagram of the intrinsic ZYX and ZXY Euler angles parameters. A 
rotation $\rotb{G}{B}{R}$ is represented by three successive elemental 
rotations, of angles $(\psiE, \thE, \phiE)$ about the $\vecs{z}{G}$, 
$\vect{y}'$, $\vecs{x}{B}$ axes (left), and of angles $(\psiEt, \phiEt, 
\thEt)$ about the $\vecs{z}{G}$, $\vect{x}'$, $\vecs{y}{B}$ axes (right), 
respectively.}
\figlabel{euler_parameters}
\vspace{-2ex}
\end{figure}
\subsection{Intrinsic ZXY Euler Angles}
\seclabel{euler_zxy}

The intrinsic ZXY Euler angles representation consists of the following three 
sequential rotations: first a rotation by the \emph{ZXY Euler yaw} $\psiEt$ 
about the z-axis, then by the \emph{ZXY Euler roll} $\phiEt$ about the new 
x-axis, and then by the \emph{ZXY Euler pitch} $\thEt$ about the newest y-axis, 
as illustrated in \figref{euler_parameters}. The complete ZXY Euler angles 
rotation is then denoted by
\begin{equation}
\rotb{G}{B}{\tilde{E}} = (\psiEt, \phiEt, \thEt) \in (-\pi,\pi] {\times} [-\hpi,\hpi] {\times} (-\pi,\pi] \equiv \Et. \eqnlabel{eulerzxydefn}
\end{equation}
Relations analogous to \eqnrefs{eulerzyxtorotmat}{eulerzyxtoquat} also hold for 
ZXY Euler angles. For instance, the quaternion $q$ corresponding to $\tilde{E}$ 
is
\begin{equation}
\begin{split}
\mspace{-6mu}q = (&c_{\phibEt} c_{\thbEt} c_{\psibEt} \!{-} s_{\phibEt} s_{\thbEt} s_{\psibEt}, s_{\phibEt} c_{\thbEt} c_{\psibEt} \!{-} c_{\phibEt} s_{\thbEt} s_{\psibEt}, \\
&c_{\phibEt} s_{\thbEt} c_{\psibEt} \!{+} s_{\phibEt} c_{\thbEt} s_{\psibEt}, c_{\phibEt} c_{\thbEt} s_{\psibEt} \!{+} s_{\phibEt} s_{\thbEt} c_{\psibEt}).\mspace{-5mu}
\end{split}
\eqnlabel{eulerzxytoquat}
\end{equation}
\vspace{-2ex}

\section{Review of Fused Angles}
\seclabel{review_fused}

We first briefly introduce the intermediate \emph{tilt angles} representation, 
and then show how the so-called \emph{tilt rotation component} is 
reparameterised to yield fused angles. More details on both representations can 
be found in \cite{Allgeuer2015a}.

\subsection{Tilt Angles}
\seclabel{tilt_angles}

Consider the rotation from a global frame \fr{G} to the body-fixed frame \fr{B}, 
as shown in \figref{tilt_parameters}. We first construct an intermediate frame 
\fr{A} by rotating $\vecs{z}{B}$ onto $\vecs{z}{G}$ in the most direct way 
possible within the plane that contains both these vectors. The \emph{fused yaw} 
$\psi$ is then defined as the angle of the z-rotation from \fr{G} to \fr{A}, and 
the \emph{tilt angle} $\alpha$ is defined as the angle of the so-called 
\emph{tilt rotation component} from \fr{A} to \fr{B}. The \emph{tilt axis angle} 
$\gamma$ defines the axis in the $\vecs{x}{G}\vecs{y}{G}$ plane about which the 
tilt rotation occurs, as shown in \figref{tilt_parameters}. The complete tilt 
angles rotation is denoted by
\begin{equation}
\rotb{G\mspace{2mu}}{B}{T} = (\psi,\gamma,\alpha) \in (-\pi,\pi] {\times} (-\pi,\pi] {\times} [0,\pi] \defeq \T. \eqnlabel{tiltdefn}
\end{equation}
All rotations with zero fused yaw $\psi$ are referred to as \emph{tilt 
rotations}, and are completely and uniquely defined by $(\gamma, \alpha)$.

\begin{figure}[!t]
\parbox{\linewidth}{\centering\includegraphics[width=0.88\linewidth]{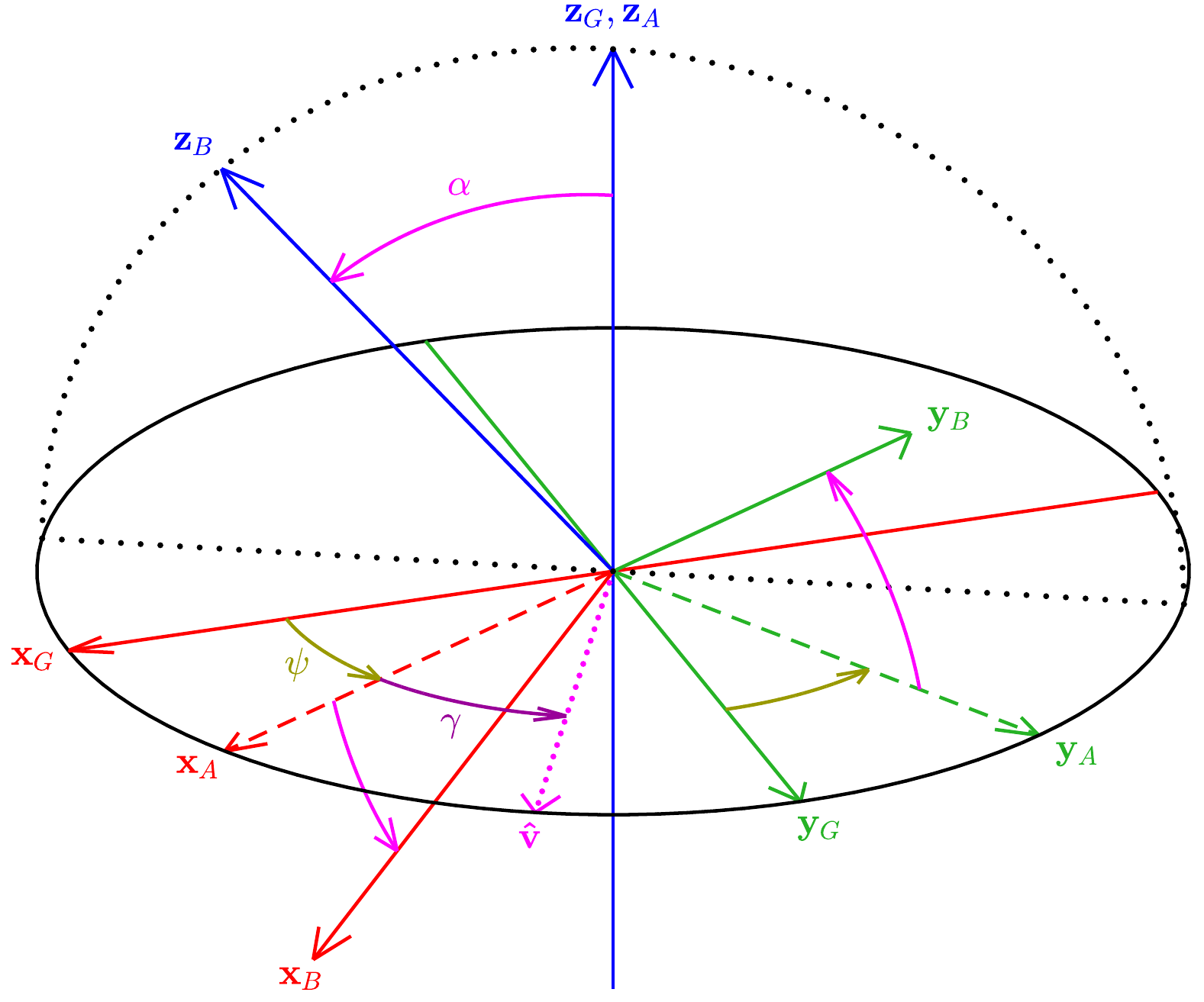}}
\caption{Diagram of the tilt angles parameters $(\psi, \gamma, \alpha)$. A 
z-rotation by $\psi$ from \fr{G} to \fr{A} is followed by a rotation by $\alpha$ 
about $\vect{\hat{v}}$ from \fr{A} to \fr{B}.}
\figlabel{tilt_parameters}
\vspace{-2ex}
\end{figure}
\subsection{Fused Angles}
\seclabel{fused_angles}

The fused angles representation also uses the fused yaw $\psi$ to represent the 
z-component of rotation, but reparameterises the tilt rotation component, as 
shown in \figref{fused_parameters}. The signed angles between $\vecs{z}{G}$ and 
the $\vecs{y}{B}\vecs{z}{B}$ and $\vecs{x}{B}\vecs{z}{B}$ planes, respectively, 
are defined as the \emph{fused pitch} $\theta$ and \emph{fused roll} $\phi$. A 
binary \emph{hemisphere} parameter $h \in \{-1,1\}$ determines which of the two 
solutions for $\vecs{z}{B}$ are taken. The complete fused angles rotation from 
\fr{G} to \fr{B} is then denoted by
\begin{equation}
\begin{aligned}
\rotb{G}{B}{F} &= (\psi,\theta,\phi,h) \\
&\in (-\pi,\pi] {\times} [-\hpi,\hpi] {\times} [-\hpi,\hpi] {\times} \{-1,1\} \defeq \hat{\F}. \eqnlabel{fuseddefn}
\end{aligned}
\end{equation}
Note that $\hat{\F}$ is used because the true domain $\F$ of fused angles is 
given by the restriction of $\hat{\F}$ by the sine sum criterion
\begin{equation}
\sin^2\theta + \sin^2\phi \leq 1 \iff \abs{\theta} + \abs{\phi} \leq \hpi. \eqnlabel{sinesumcriterion}
\end{equation}
Note that $\psi$, $\theta$ and $\phi$ quantify the amount of rotation within the 
$\vect{xy}$, $\vect{xz}$ and $\vect{yz}$ major planes of \fr{A}, respectively. 
Mathematically, if $q = (w,x,y,z) \in \Q$ is the corresponding quaternion, the 
fused angle parameters are given by
\begin{equation}
\begin{aligned}
\psi &= \wrap\bigl(2\atantwo(z,w)\bigr), & \theta &= \asin\bigl(2(wy-xz)\bigr), \\
h &= \sign(w^2+z^2-\half), & \phi &= \asin\bigl(2(wx+yz)\bigr),
\end{aligned}
\eqnlabel{fusedfromquat}
\end{equation}
where $\wrap(\cdot)$ is a function that wraps an angle to $(-\pi,\pi]$. If $R$ 
is the corresponding rotation matrix, and $R_{ij}$ denotes the matrix entries, 
the fused pitch and roll are also given by
\begin{equation}
\begin{aligned}
\theta &= \asin(-R_{31}), & \quad \phi &= \asin(R_{32}).
\end{aligned}
\eqnlabel{fusedfromrotmat}
\end{equation}
The rotation matrix for $T = (\psi, \gamma, \alpha)$, $F = (\psi, \theta, \phi, 
h)$ and $\delta \equiv \psi + \gamma$ is given by
\begin{equation}
R = 
\begin{bmatrix}
c_\gamma c_\delta + c_\alpha s_\gamma s_\delta & s_\gamma c_\delta - c_\alpha c_\gamma s_\delta & s_\alpha s_\delta \\
c_\gamma s_\delta - c_\alpha s_\gamma c_\delta & s_\gamma s_\delta + c_\alpha c_\gamma c_\delta & -s_\alpha c_\delta \\
-s_\theta & s_\phi & c_\alpha
\end{bmatrix}\mspace{-6mu}.
\eqnlabel{tiltfusedtorotmat}
\end{equation}

\begin{figure}[!t]
\parbox{\linewidth}{\centering\includegraphics[width=0.88\linewidth]{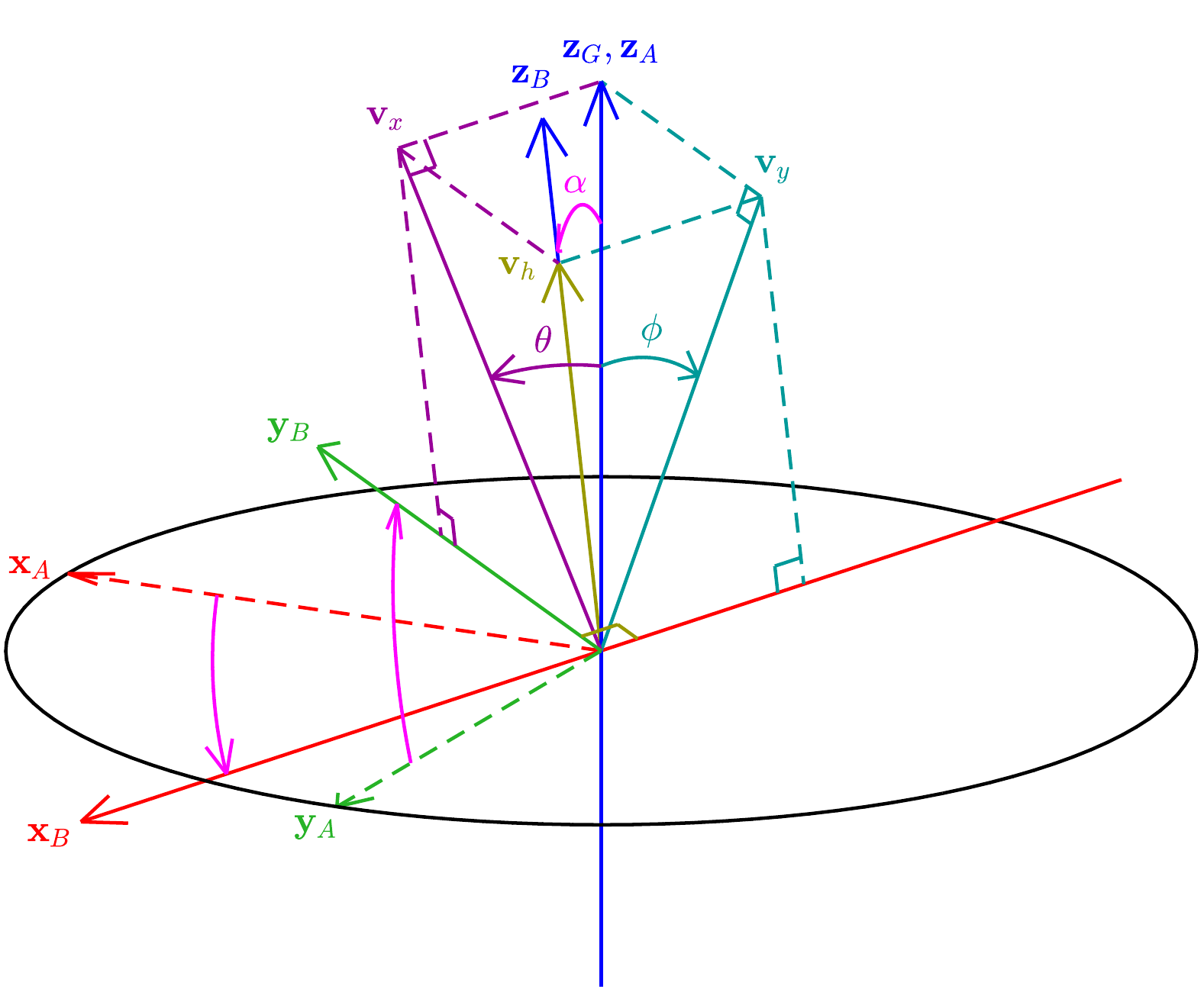}}
\caption{Diagram of the fused angles parameters $(\theta, \phi, h)$. The 
rotation by $\alpha$ from \fr{A} to \fr{B} is reparameterised by the angles 
$\theta$ and $\phi$, defined between $\vecs{z}{G}$ and the 
$\vecs{y}{B}\vecs{z}{B}$ and $\vecs{x}{B}\vecs{z}{B}$ planes, respectively. The 
hemisphere $h$ is $1$ if $\vecs{v}{h}$ is parallel to $\vecs{z}{B}$, and $-1$ if 
it is antiparallel.}
\figlabel{fused_parameters}
\vspace{-2ex}
\end{figure}
\section{Fundamental Properties and Results}
\seclabel{properties_results}

The following properties and results are required for the comparative analysis 
between fused angles and Euler angles.

\subsection{Mathematical Links Between Fused and Euler Angles}
\seclabel{links_fused_euler}

Even though the interpretations of the variables are quite different, and the 
nature of the domains do not correspond, purely mathematically it can be 
observed that
\begin{equation}
\begin{aligned}
\thE &= \theta, & \quad \phiEt &= \phi.
\end{aligned}
\eqnlabel{fusedeulerequalities}
\end{equation}
As such, fused angles can be seen to---with an adaptation of the domains and 
geometric interpretation---unite the ZYX Euler pitch and ZXY Euler roll with a 
novel and meaningful concept of yaw, to form a useful representation for 
rotations.

The following equations relate the ZYX Euler angles, fused angles and tilt 
angles parameters:
\begin{gather}
\begin{aligned}
\phiE &= \atantwo(s_\phi, c_\alpha), & \quad \gamma &= \atantwo(s_{\thE}, c_{\thE} s_{\phiE}), \\
\phi &= \asin(c_{\thE} s_{\phiE}), & \quad \alpha &= \acos(c_{\thE} c_{\phiE}),
\end{aligned} \eqnlabel{fusedeulerrelA} \\
\mspace{-24mu}h = \sign(c_{\phiE}) =
\begin{cases}
1, & \text{if $\abs{\phiE} \leq \hpi$,} \\
-1, & \text{otherwise,}
\end{cases} \eqnlabel{fusedeulerrelB} \\
\mspace{-24mu}s_\alpha^2 = s_{\thE}^2 + s_{\phiE}^2 - s_{\thE}^2 s_{\phiE}^2. \eqnlabel{fusedeulerrelC}
\end{gather}
Away from the fused yaw and Euler yaw singularities, the relationship between 
the two yaws is given by
\begin{align}
\psiE &= \wrap(\psi + \gamma - \atantwo(c_\alpha s_\gamma, c_\gamma)), \eqnlabel{fusedeuleryawrelations} \\
\psi &= \wrap(\psiE - \atantwo(s_\theta, s_{\phi}) + \atantwo(s_\theta c_{\phiE}, s_{\phiE}) \notag\\
&= \wrap(\psiE - \atantwo(s_{\thE}, c_{\thE} s_{\phiE}\mspace{-2mu}) + \atantwo(s_{\thE} c_{\phiE}, s_{\phiE}\mspace{-2mu})\mspace{-1mu}. \notag
\end{align}

\subsection{Effect of Pure Z-Rotations on the Fused Yaw}
\seclabel{effect_purez_yaw}

Unlike for Euler yaw, the composition of any rotation with a pure z-rotation is 
additive in terms of fused yaw, irrespective of whether the z-rotation is local 
or global, i.e.\ applied by post-multiplication or pre-multiplication. If 
$\Psi(\cdot)$ is the generic operator that returns the fused yaw of a rotation 
in any representation, then up to angle wrapping
\begin{equation}
\yawof{R R_z(\psi_z)} = \yawof{R_z(\psi_z) R} = \Psi(R) + \psi_z, \eqnlabel{purezadditivefusedyaw}
\end{equation}
where $R_z(\psi_z)$ is the rotation matrix corresponding to a pure z-rotation by 
$\psi_z$. For global z-rotations, the tilt rotation component also remains 
unchanged. That is,
\begin{equation}
\begin{aligned}
T_z(\psi_z) \circ T(\psi, \gamma, \alpha) &= T(\psi + \psi_z, \gamma, \alpha), \\
F_z(\psi_z) \circ F(\psi, \theta, \phi, h) &= F(\psi + \psi_z, \theta, \phi, h),
\end{aligned}
\eqnlabel{purezpremult}
\end{equation}
where $T(\cdot)$, $F(\cdot)$ is notation that clarifies that the enclosed 
parameters are tilt angles or fused angles, respectively, and $T_z(\psi_z)$, 
$F_z(\psi_z)$ correspond to pure z-rotations by $\psi_z$.

\subsection{Format of Rotation Inverses}
\seclabel{format_inverses}

The inverses of quaternion rotations and rotation matrices are simply given by:
\begin{equation}
\begin{aligned}
q\inv &= q\conj, & \quad R\inv &= R\trans,
\end{aligned}
\eqnlabel{invquatrotmat}
\end{equation}
where $q\conj$ is the quaternion conjugate. For Euler angles, the situation is 
more complicated. For $E = (\psiE, \thE, \phiE)$,
\begin{equation}
\begin{gathered}
E\inv = (\sub{\psi}{Einv}, \sub{\theta}{Einv}, \sub{\phi}{Einv}), \\
\begin{aligned}
\sub{\psi}{Einv} &= \atantwo(c_{\psiE} s_{\thE} s_{\phiE} - s_{\psiE} c_{\phiE}, c_{\psiE} c_{\thE}), \\
\sub{\theta}{Einv} &= -\asin(c_{\psiE} s_{\thE} c_{\phiE} + s_{\psiE} s_{\phiE}), \\
\sub{\phi}{Einv} &= \atantwo(s_{\psiE} s_{\thE} c_{\phiE} - c_{\psiE} s_{\phiE}, c_{\thE} c_{\phiE}).
\end{aligned}
\end{gathered}
\eqnlabel{inveuler}
\end{equation}
For tilt and fused angles, the rotation inverses are given by
\begin{equation}
\begin{gathered}
\begin{aligned}
T\inv &= (-\psi, \wrap(\psi + \gamma - \pi), \alpha), \\
F\inv &= (-\psi, \sub{\theta}{inv}, \sub{\phi}{inv}, h),
\end{aligned} \\
\begin{aligned}
\sub{\theta}{inv} &= -\asin(s_\alpha s_{\psi + \gamma}) = -\asin(c_\psi s_\theta + s_\psi s_\phi), \\
\sub{\phi}{inv} &= -\asin(s_\alpha c_{\psi + \gamma}) = \asin(s_\psi s_\theta - c_\psi s_\phi).
\end{aligned}
\end{gathered}
\eqnlabel{invtiltfused}
\end{equation}
It is quite remarkable to note from \eqnref{invtiltfused} that
\begin{equation}
\yawof{R\inv} = -\yawof{R}. \eqnlabel{invfusedyaw}
\end{equation}
This property of fused yaw is referred to as negation through rotation 
inversion, and is clearly not satisfied by any variant of Euler yaw. 
Furthermore, for the case of pure tilt rotations, i.e.\ zero fused yaw, the 
fused pitch and roll also satisfy the negation through rotation inversion 
property:
\begin{equation}
\psi = 0 \iff F\inv = (0, -\theta, -\phi, h). \eqnlabel{invnoyawfusedpitchroll}
\end{equation}
For zero Euler yaw $\psiE$, the expression for the inverse rotation does not 
simplify as significantly:
\begin{equation}
\psiE = 0 \iff
\begin{aligned}
E\inv = \bigl( & \atantwo(s_{\thE} s_{\phiE}, c_{\thE}), \\
&\asin(-s_{\thE} c_{\phiE}), \\
&\atantwo(-s_{\phiE}, c_{\thE} c_{\phiE}) \bigr).
\end{aligned}
\eqnlabel{invnoyaweuler}
\end{equation}

\section{Comparative Analysis}
\seclabel{comparative_analysis}

The fused and Euler angles representations are critically compared in this 
section. In particular, the many differences between the two representations 
that make fused angles superior to Euler angles for representing orientations 
are delineated. The main drawbacks of Euler angles are:
\begin{enumerate}[label=\Alph*)]
\item The proximity of the gimbal lock singularity to normal working ranges, 
leading to unwanted artefacts due to increased local parameter sensitivities in 
a widened neighbourhood of the singularity,
\item The interdependence of the Euler parameters, leading to an unclear 
attribution of which parameter encapsulates which major plane of rotation,
\item The asymmetry introduced by the use of a definition of yaw that depends on 
projection, leading to unintuitive non-axisymmetric behaviour of the yaw angle, 
and
\item The fundamental requirement of an order of elemental rotations, leading to 
non-axisymmetric definitions of pitch and roll that do not correspond in 
behaviour.
\end{enumerate}

\subsection{Singularities and Local Parameter Sensitivities}
\seclabel{singularities}

It was shown by Stuelpnagel \cite{Stuelpnagel1964} that it is topologically 
impossible to have a global three-dimensional parameterisation of the rotation 
group without any singular points. That is, every three-dimensional 
parameterisation of the rotation space, including both Euler angles and fused 
angles, must have at least one of the following:
\begin{enumerate}[label=(\roman*)]
\item \enumlabel{singularity_type_1} A rotation that does not have a unique set 
of parameters,
\item \enumlabel{singularity_type_2} A parameter set that does not specify a 
unique rotation,
\item \enumlabel{singularity_type_3} A rotation in the neighbourhood of which 
the sensitivity of the map from rotations to parameters is unbounded.
\end{enumerate}

The Euler angles representation is singular at gimbal lock, i.e.\ $\thE = 
\pm\hpi$. For $\lambda \in \R$, the following equivalences hold:
\begin{equation}
\begin{aligned}
(\psiE, \hpi, \phiE) &\equiv (\psiE - \lambda, \hpi, \phiE - \lambda), \\
(\psiE, -\hpi, \phiE) &\equiv (\psiE - \lambda, -\hpi, \phiE + \lambda).
\end{aligned}
\eqnlabel{euler_singularity_equiv}
\end{equation}
It can be seen that $\psiE$, $\phiE$ both have essential discontinuities at 
gimbal lock, and each correspond to type \enumref{singularity_type_1} and 
\enumref{singularity_type_3} singularities. Fused angles only possess a single 
singularity:
\begin{equation}
\begin{aligned}
\text{Singular $\psi$} &\iff \alpha = \pi \iff w = z = 0 \\
&\iff \theta = \phi = 0\ \text{and}\ h = -1 \\
&\iff R_{33} \equiv \compbs{G}{z}{Bz} \equiv \compbs{B}{z}{Gz} = -1.
\end{aligned}
\end{equation}
The so-called fused yaw singularity is also an essential discontinuity, and is 
of type \enumref{singularity_type_1} and \enumref{singularity_type_3}, when 
considering the geometric definition of the parameters, like for Euler angles.

The local state of balance of a body is a function of pitch and roll, but not 
yaw, as this just determines the heading. Thus, it is critical to compare that 
fused angles have a single singularity in a single parameter, namely the fused 
yaw, while Euler angles have two singularities in two parameters, including, 
very importantly, one that is not yaw. Hence, fused angles can represent local 
states of balance completely without singularities, while this is not the case 
for Euler angles. The fused yaw singularity is also `maximally far' from the 
identity rotation, at 180\degree, while the two Euler singularities are only 
90\degree away, which is close to, if not in, normal working ranges. In fact, 
the increased parameter sensitivity of the Euler yaw and roll near gimbal lock 
has noticeable effects even for tilt rotations of only 65\degree, as can be seen 
in \figref{sensitivity}. Sudden sensitive changes in Euler yaw and roll occur 
even when the tilt rotation is actually only a few degrees from being pure 
pitch---something that is highly problematic. Consequently, the Euler yaw 
component of a rotation cannot in general be meaningfully removed, as for even 
moderate tilts this can lead to large z-rotations occurring in the rotation that 
remains, which should actually only be the contribution of pitch and roll.

\begin{figure}[!t]
\parbox{\linewidth}{\centering\includegraphics[width=1.00\linewidth]{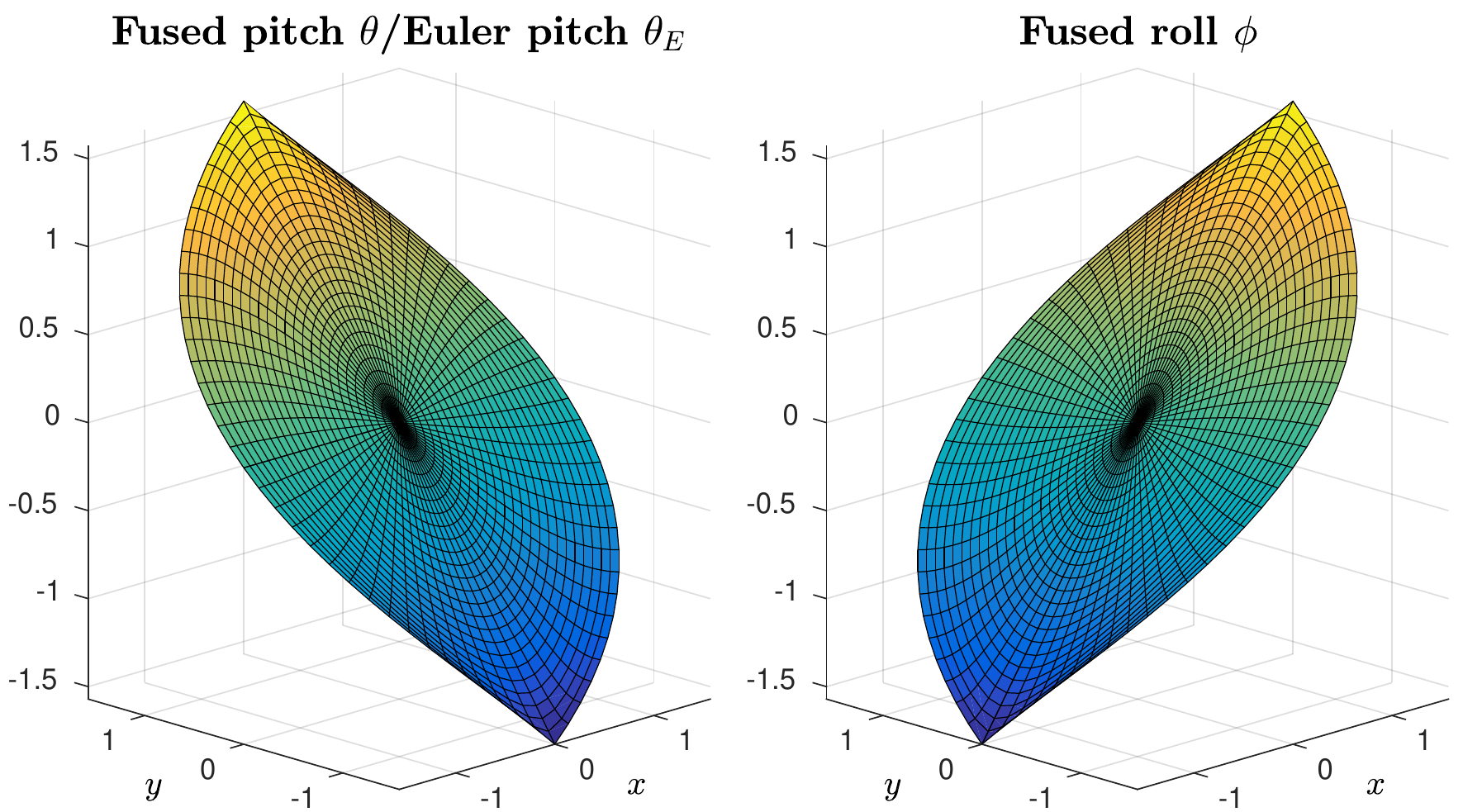}\\[1ex]
\includegraphics[width=1.00\linewidth]{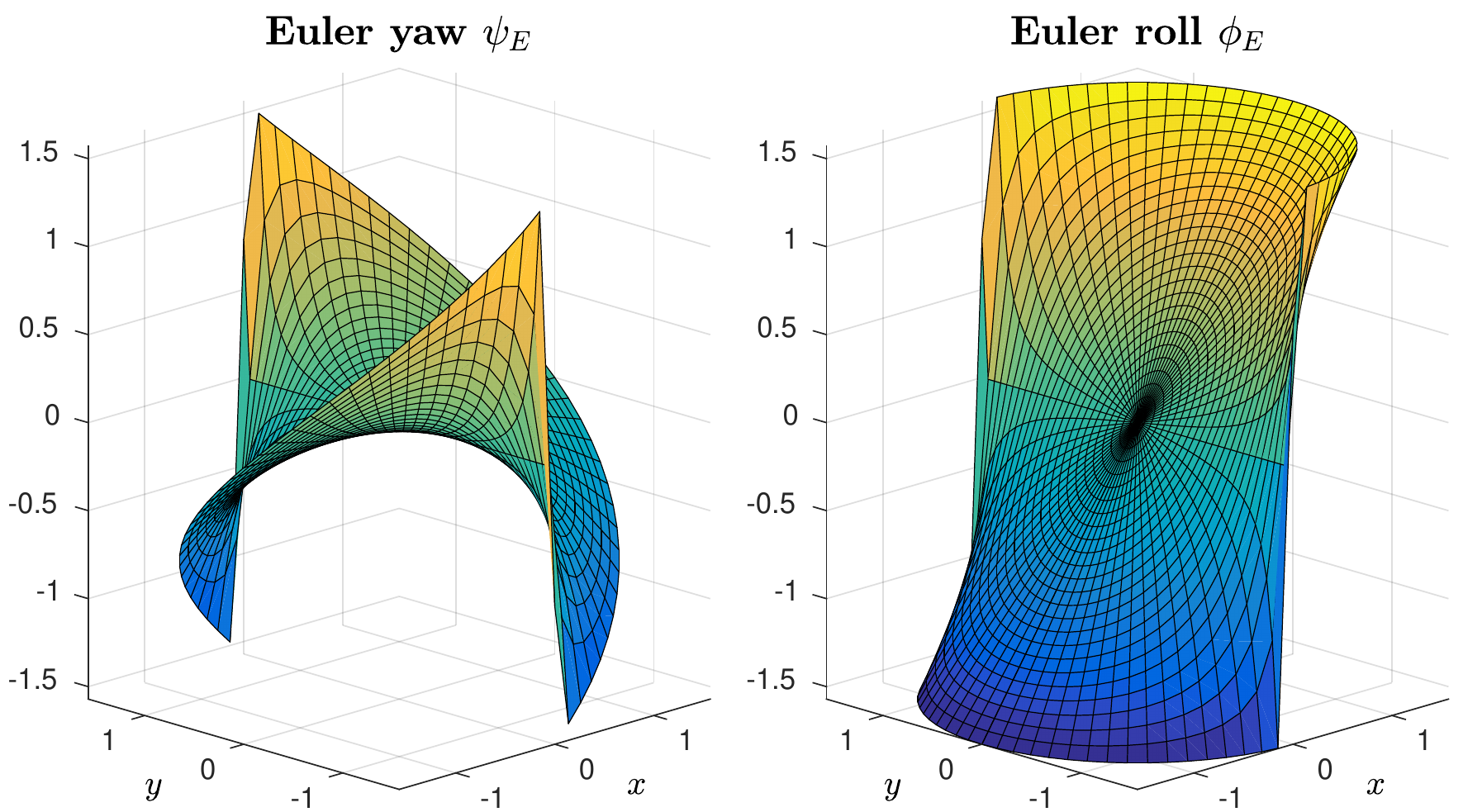}}
\caption{Plots of the fused and Euler angles parameters when an upright body is 
rotated about every axis in the $\vect{x}\vect{y}$ plane. These are pure tilt 
rotations as no component of rotation is about the z-axis. The parameters of a 
rotation by $\alpha$ about the axis $(c_\gamma, s_\gamma, 0)$ are plotted at 
$(x,y) = (\alpha c_\gamma, \alpha s_\gamma)$. The fused yaw plot is omitted 
because it is simply perfectly zero everywhere. It can clearly be seen that 
fused pitch and roll correspond to each other in behaviour, while Euler pitch 
and roll do not, and that Euler yaw is not axisymmetric, while the fused yaw 
trivially is, as it is zero everywhere.}
\figlabel{sensitivity}
\vspace{-2ex}
\end{figure}
\subsection{Mutual Independence of Rotation Parameters}
\seclabel{mutual_independence}

To fulfil the parameterisation aims that were set out in \secref{introduction}, 
one necessary condition is that the individual parameters should be as mutually 
independent as possible, and correspond intuitively to the x, y and z-components 
of rotation. This is not the case for Euler angles, as is shown by comparison to 
fused angles in the following subsections.

\textbf{Interdependence of Yaw and Roll:} After the Euler yaw elemental rotation 
has been applied, the axis $\vect{y}'$ of the following Euler pitch rotation 
(see \figref{euler_parameters}) always lies in the $\vect{x}\vect{y}$ plane. For 
the Euler roll rotation however, the axis $\vect{x_B}$ in general has a non-zero 
z-component, and thus applies a contribution to `yaw' in the intuitive sense. 
This effect can be seen most clearly at gimbal lock, where Euler yaw and roll 
become completely interchangeable, as given by \eqnref{euler_singularity_equiv}. 
Conceptually, part of the total `yaw' of a rotation is always quantified by the 
$\phiE$ parameter instead of $\psiE$, meaning that neither parameter cleanly 
represents the component of rotation that it ideally should. This can also be 
observed in \figref{sensitivity}. Fused angles do not have this kind of 
interdependence.

\textbf{Interdependence of Pitch and Roll:} As the Euler pitch elemental 
rotation precedes the Euler roll one, the axis of rotation $\vect{x_B}$ of the 
latter is a function of $\thE$. This creates a dependency of $\phiE$ on $\thE$, 
which results in $\phiE$ not completely capturing the intuitive sense of `roll' 
independently by itself. This can be seen in the bottom row of 
\eqnref{eulerzyxtorotmat}, which is a heading-independent measure of the global 
up direction, i.e.\ just like an accelerometer would measure gravity. While the 
$R_{31}$ entry is a pure function of $\thE$, the $R_{32}$ entry is not a 
function purely of $\phiE$, as would naturally be desired. It can be seen from 
\eqnref{tiltfusedtorotmat} however, that both these properties hold for fused 
angles.

\textbf{Purity of the Axis of Rotation:}
Euler's rotation theorem \cite{Palais2007} states that every rotation can be 
expressed as a single rotation about some vector $\vect{e}$. By definition, this 
vector must lie on the line defined by the $(x,y,z)$ quaternion parameters. 
Thus, the relative ratios of these quaternion parameters gives insight into the 
proportions of the rotation that are about each of the corresponding axes. It is 
known from \eqnref{fusedfromquat} that
\begin{equation}
\psi = 0 \iff z = 0. \eqnlabel{zerofusedyawquat}
\end{equation}
This can be interpreted as saying that the fused yaw is zero exactly when there 
is no component of rotation about the z-axis. This is quite logical, but not the 
case for Euler yaw.

The yaw components $\psi$ and $\psiE$ simply rotate the axis of rotation 
$\vect{e}$ around the z-axis by half their value. As such, we can inspect the 
purity of the x and y-axis components by examining just rotations with zero yaw. 
For fused angles with zero $\psi$, it can be deduced from \eqnref{fusedfromquat} 
that $\vect{e}$ is on the line $(s_\phi, s_\theta, 0)$. From 
\eqnref{eulerzyxtoquat}, for Euler angles with zero $\psiE$, $\vect{e}$ is on 
the line $(s_{\phibE} c_{\thbE}, c_{\phibE} s_{\thbE}, -s_{\phibE} s_{\thbE})$. 
While for fused angles it can be seen that there is no component of rotation 
about the x and y-axes exactly when $\phi$ and $\theta$ are zero respectively, 
for Euler angles, $e_y$ is also zero when $\phiE = \pi$. This comes about 
because the $e_x$ and $e_y$ components are mixed expressions of Euler pitch and 
roll, instead of clean independent expressions like for fused angles, where 
direct one-to-one associations can be made between $e_x\!\leftrightarrow\!\phi$ 
and $e_y\!\leftrightarrow\!\theta$. It is also evident from the $e_z = 
-s_{\phibE} s_{\thbE}$ term that Euler pitch and roll together contribute a 
component of rotation about the z-axis, which is unintuitive. In fact, from 
\eqnref{eulerzxytoquat}, the ZXY Euler angles expression for $\vect{e}$ is 
$(s_{\phibEt} c_{\thbEt}, c_{\phibEt} s_{\thbEt}, s_{\phibEt} s_{\thbEt})$, 
which has the exact opposite contribution to $e_z$. As such, as $e_z = 0$, fused 
angles can conceptually be thought of as being exactly inbetween ZYX and ZXY 
Euler angles in terms of how the x and y contributions are combined---concurrent 
and neutral, instead of asymmetrical due to a discrete order of rotations.

\textbf{Rotation Inverses:}
The fused yaw satisfies the remarkable negation through rotation inversion 
property, \eqnref{invfusedyaw}. This property is quite logical, as the component 
of rotation about the z-axis is negated for the inverse rotation, as can be seen 
from the inverse quaternion in \eqnref{invquatrotmat}. As a corollary, the 
inverse of a zero fused yaw rotation also has zero fused yaw. Despite being very 
natural, neither of these two properties hold for Euler yaw. The inverse 
equation \eqnref{inveuler} for Euler yaw actually depends on both Euler pitch 
and roll, demonstrating that these parameters are interdependent. Even for 
rotations with zero Euler yaw, \eqnref{invnoyaweuler} shows that all inverse 
terms are mixed combinations of pitch and roll, including notably the non-zero 
inverse Euler yaw. By comparison, it can be seen from 
\eqnref{invnoyawfusedpitchroll} that for rotations with zero fused yaw, the 
inverse fused angles rotation resolves trivially into negation through rotation 
inversion for both fused pitch and roll.

\subsection{Axisymmetry of Yaw}
\seclabel{axisymmetry_yaw}

When using the fused angles representation in balance-related applications, by 
design the z-axis is chosen to point in the direction opposite to gravity. This 
ensures that the concepts of roll, pitch, and in particular yaw, line up with 
what one would intuitively expect. The choice of z-axis however still leaves one 
degree of rotational freedom open for the choice of global x and y-axis. In the 
context of this paper, the concept of \emph{axisymmetry} refers to the property 
that one or more rotation parameters are either invariant to this freedom of 
choice in the axes, or vary in an intuitive rotational manner proportional to 
the choice. In other words, axisymmetry refers to the notion that the rotation 
parameters, in order to be self-consistent, should be symmetrical about the 
unambiguously defined z-axis. This is a relatively natural property to desire, 
as, for example, the amount of yaw a rotation has should clearly transcend any 
arbitrary choice of which reference frame to use.

\begin{figure}[!t]
\parbox{\linewidth}{\centering\includegraphics[width=1.0\linewidth]{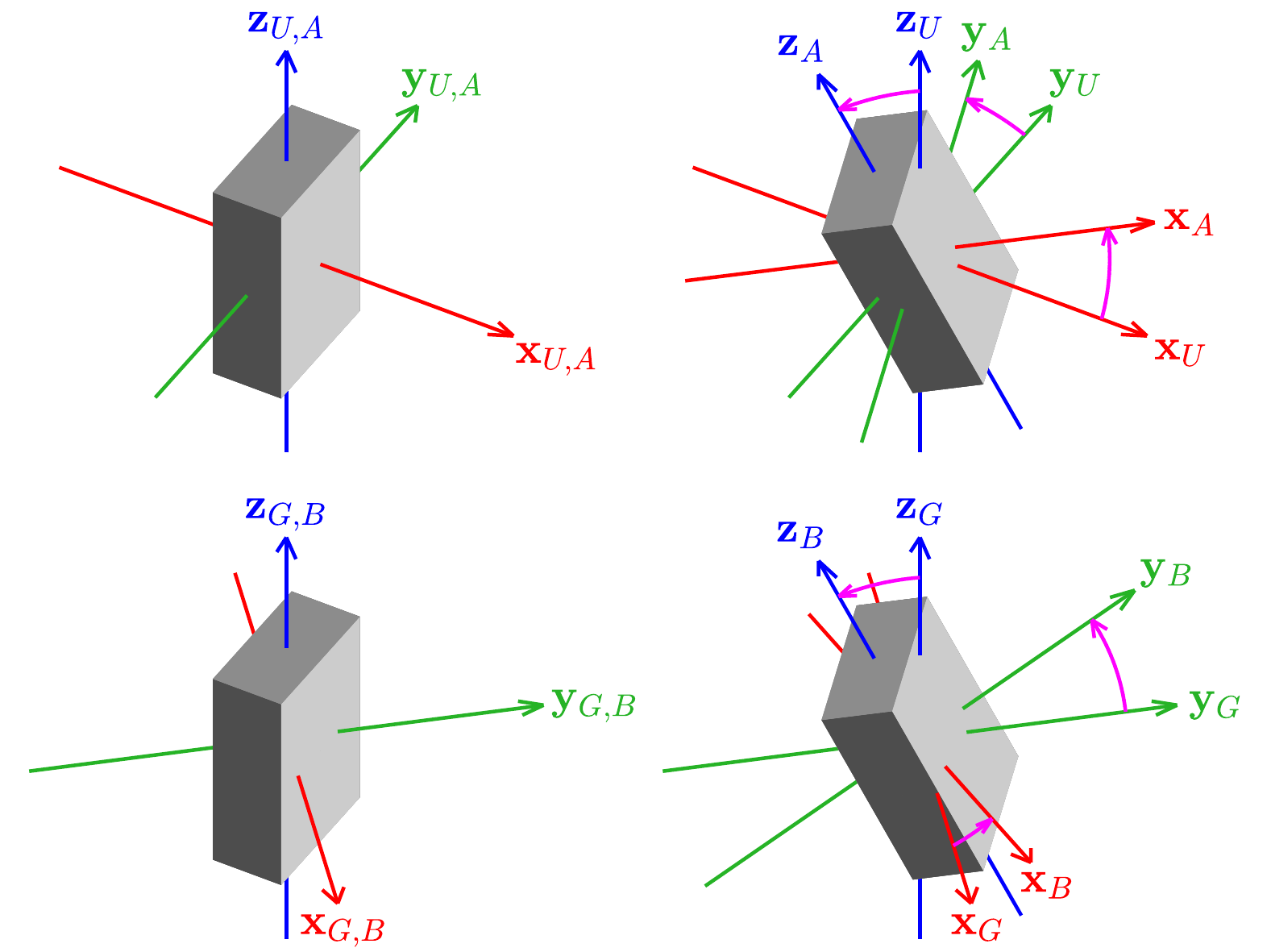}}
\caption{Definition of frames for the investigation of axisymmetry, where the 
left side in each row is before rotation, and the right side is after. The same 
physical rotation is applied in each row. Prior to rotation, \fr{U} = \fr{A} and 
\fr{G} = \fr{B}, but \fr{U} and \fr{G} are fixed to the environment, while 
\fr{A} and \fr{B} rotate with the robot. Thus, $\rotb{U}{A}{R}$ and 
$\rotb{G}{B}{R}$ represent the exact same physical rotation, but with respect to 
different reference frames, and so are not numerically equal. The axisymmetry of 
fused yaw asserts that irrespective of the choice of \fr{G}, the fused yaws of 
$\rotb{U}{A}{R}$ and $\rotb{G}{B}{R}$ are equal.}
\figlabel{axisymmetry_rotation}
\vspace{-2ex}
\end{figure}

The fused yaw is axisymmetric in the sense that it is invariant to the choice of 
global x and y-axis. Consider a robot that is upright, and thereby considered 
mathematically to have an identity orientation $\I_3$ relative to its 
environment. If the robot undergoes any rotation, the above statement of fused 
yaw axisymmetry asserts that the fused yaw of this rotation is the same no 
matter what choice of global x and y-axis was made. This is an important and 
reassuring property of the fused yaw as, given that the z-axis is unambiguously 
defined, any concept of yaw about the z-axis should clearly be a property of the 
actual physical rotation, not a property of some arbitrary choice of virtual 
reference frame made solely for the purpose of mathematical analysis. It can 
easily be demonstrated, with virtually any non-degenerate example, that Euler 
yaw is not axisymmetric, and for different choices of axes can readily produce 
deviations up to 180\degree.

Let \fr{U} be a global coordinate frame such that $\vecs{z}{U}$ points in the 
direction opposite to gravity, as required, and suppose that the rotation that 
is undergone by the robot is given by $\rotb{U}{A}{R}$. This is a fixed physical 
rotation of the robot relative to its environment, so it should have a unique 
well-defined fused yaw according to axisymmetry. As the z-axis is fixed, every 
valid global coordinate system \fr{G} that can be used as a reference frame to 
quantify $\rotb{U}{A}{R}$, including \fr{U} itself, is a pure z-rotation of 
\fr{U}. That is, for some angle $\beta$,
\begin{equation}
\rotb{U}{G}{R} = \rots{R}{z}(\beta). \eqnlabel{UGRpurez}
\end{equation}
Given any choice of \fr{G}, a frame \fr{B} is attached to the robot in such a 
way that it coincides with \fr{G} when the robot is initially upright, and 
rotates with the robot, as shown in \figref{axisymmetry_rotation}. As such, 
$\rotb{U}{A}{R}$ and $\rotb{G}{B}{R}$ are simply two different ways of 
quantifying the exact same rotation, just with a different frame of reference. 
The rotation $\rotb{U}{A}{R}$ maps \fr{G} onto \fr{B}, so
\begin{align}
\rotb{U}{A}{R} = \rotb{U}{G}{R} \, \rotb{G}{B}{R} \, \rotb{G}{U}{R}. \eqnlabel{UARexpansion}
\end{align}
Taking the fused yaw of both sides, twice applying 
\eqnref{purezadditivefusedyaw} as $\rotb{U}{G}{R}$ and $\rotb{G}{U}{R}$ are both 
pure z-rotations, and using \eqnref{invfusedyaw}, gives
\begin{align*}
\yawof{\rotb{U}{A}{R}} &= \yawof{\rotb{U}{G}{R} \, \rotb{G}{B}{R} \, \rotb{G}{U}{R}} \\
&= \yawof{\rotb{U}{G}{R}} + \yawof{\rotb{G}{B}{R}} + \yawof{\rotb{U}{G}{R}\trans\,} \\
&= \yawof{\rotb{U}{G}{R}} + \yawof{\rotb{G}{B}{R}} - \yawof{\rotb{U}{G}{R}} \\
&= \yawof{\rotb{G}{B}{R}}.
\end{align*}
We note that $\yawof{\rotb{U}{A}{R}}$ is clearly independent of the choice of 
\fr{G}, so $\yawof{\rotb{G}{B}{R}}$ must also be. This demonstrates that the 
fused yaw of the rotation is invariant to the choice of reference x and y-axis, 
i.e.\ choice of $\beta$, as required.

\begin{figure}[!t]
\parbox{\linewidth}{\centering\includegraphics[width=1.0\linewidth]{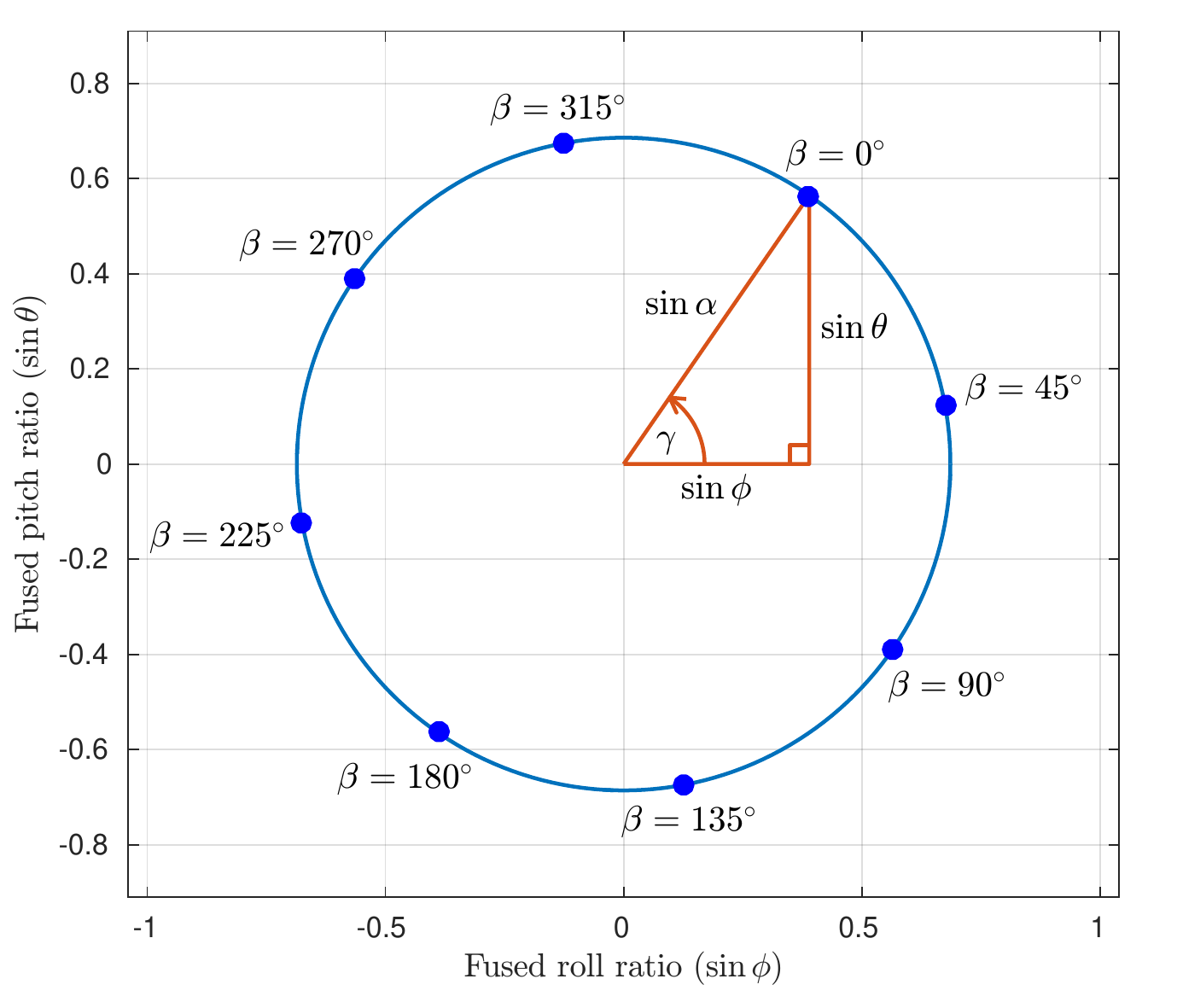}}
\caption{Locus of the sine ratios $(\sin\phi_\beta, \sin\theta_\beta)$ as 
$\beta$ varies, i.e.\ for all possible choices of x and y-axis, where 
$(\theta_0, \phi_0) = (0.6, 0.4)$. Refer to \eqnref{UAGGBRfusedtilt} for the 
definitions of $\theta_\beta$, $\phi_\beta$, $\theta_0$ and $\phi_0$. The 
triangle demonstrates the decomposition of $\sin\alpha$ into the quadrature 
sinusoid components $\sin\theta$ and $\sin\phi$, and shows how $\beta$ can be 
seen as a negative offset to $\gamma$.}
\figlabel{pitch_roll_beta}
\vspace{-2ex}
\end{figure}

To show that Euler yaw violates axisymmetry, consider
\begin{equation*}
\begin{aligned}
\rotb{U}{A}{R} &= \rots{R}{x}(\tfrac{3\pi}{4}), &\quad \beta &= \hpi.
\end{aligned}
\end{equation*}
The Euler yaw of $\rotb{U}{A}{R}$ is clearly zero, but from 
\eqnrefs{UGRpurez}{UARexpansion},
\begin{align}
\rotb{G}{B}{R} &= \rotb{G}{U}{R} \, \rotb{U}{A}{R} \, \rotb{U}{G}{R} \eqnlabel{GBRexpansion} \\
&= \rots{R}{z}(-\hpi) \rots{R}{x}(\tfrac{3\pi}{4}) \rots{R}{z}(\hpi) \notag \\
&= \sub{E}{R}(\pi, -\tfrac{\pi}{4}, \pi), \notag
\end{align}
where $\sub{E}{R}(\cdot)$ denotes the rotation matrix corresponding to the given 
Euler angles parameters. Thus, the Euler yaw of $\rotb{G}{B}{R}$ is $\pi$, which 
is totally different to that of $\rotb{U}{A}{R}$. This proves that the Euler yaw 
cannot be axisymmetric. The non-axisymmetry of Euler yaw is visualised in 
\figref{euler_param_beta}.

\subsection{Axisymmetry of Pitch and Roll}
\seclabel{axisymmetry_pitch_roll}

The fused pitch and roll are axisymmetric in the sense that their \emph{sine 
ratios} $\sin\theta$, $\sin\phi$ circumscribe a uniform circle as a function of 
the choice of x and y-axis. That is, the locus of $(\sin\phi, \sin\theta)$ over 
all possible choices of axes is a circle, and this circle is traversed uniformly 
as the choice varies. As demonstrated later, this is not the case for Euler 
angles.

\begin{figure}[!t]
\parbox{\linewidth}{\centering\includegraphics[width=1.0\linewidth]{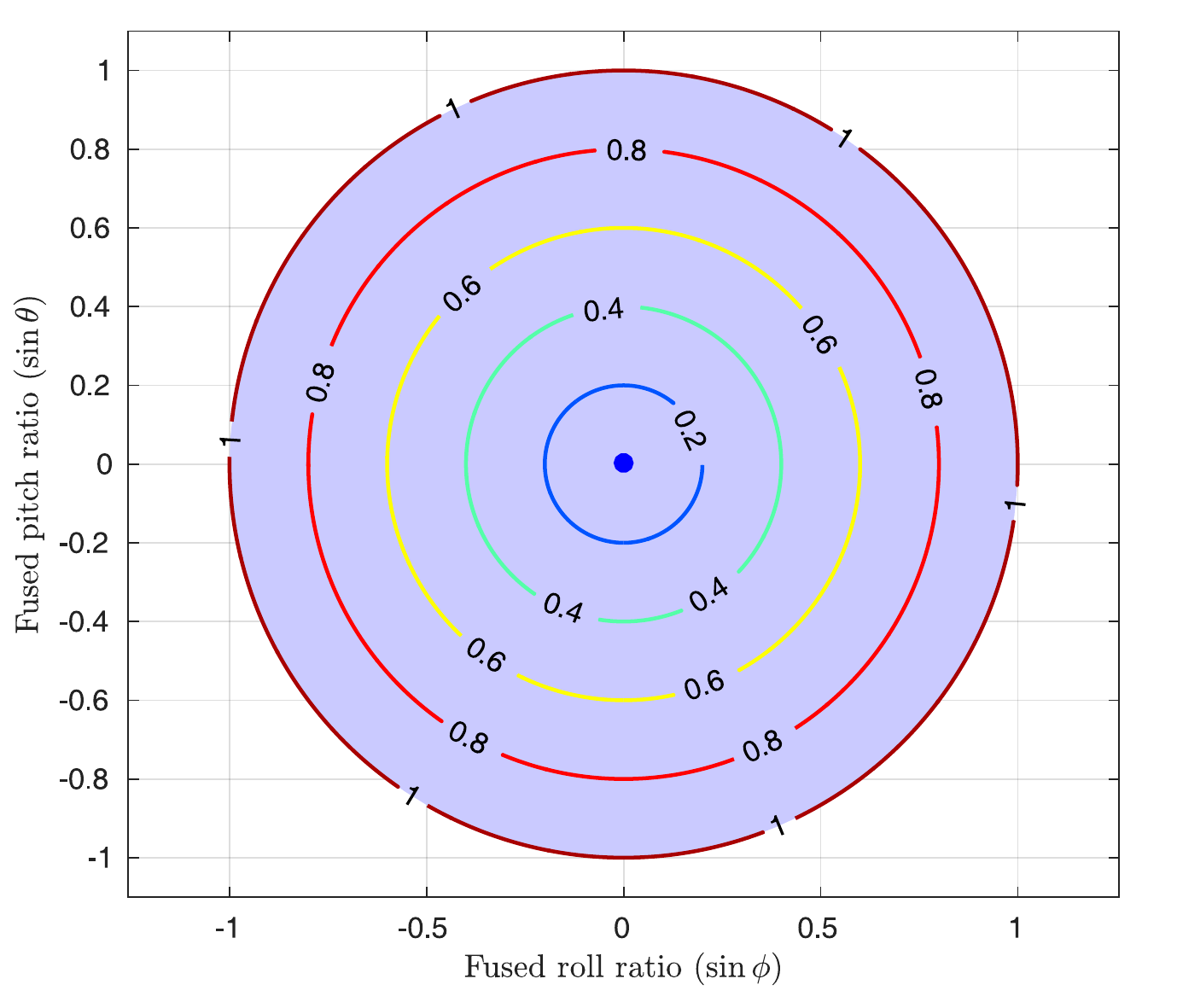}}
\caption{Level sets of constant $\sin\alpha$---the sine of the magnitude of the 
tilt rotation component of a rotation---in the fused pitch ratio $\sin\theta$ 
vs.\ fused roll ratio $\sin\phi$ Cartesian space. The shaded region is the valid 
domain of $(\sin\phi, \sin\theta)$ for the fused angles representation. The 
purely circular nature of the plot visually illustrates the axisymmetry of fused 
pitch and roll.}
\figlabel{pitch_roll_axisymmetry}
\vspace{-2ex}
\end{figure}

The fused pitch $\theta$ and fused roll $\phi$ come together with the hemisphere 
$h$ to define the tilt rotation component of a rotation. The magnitude of this 
tilt rotation is given by the tilt angle $\alpha$, and the relative direction of 
this tilt rotation is given by the tilt axis angle $\gamma$. The angles $\phi$ and $\theta$ 
can be thought of as a way of `splitting up' the action of $\alpha$ into its 
orthogonal components. More precisely, the sine ratios $\sin\phi$ and 
$\sin\theta$ are in fact a decomposition of $\sin\alpha$ into quadrature 
sinusoid components, as illustrated in \figref{pitch_roll_beta} and 
\figref{euler_param_beta}, and as embodied by
\begin{equation}
\begin{aligned}
\sin\alpha &= \sqrt{\sin^2\phi + \sin^2\theta}, \\
\gamma &= \atantwo\bigl(\sin\theta,\sin\phi\bigr).
\end{aligned}
\eqnlabel{fusedpitchrollquadrature}
\end{equation}
The property of axisymmetry in fused pitch and roll is equivalent to stating 
that the choice of global x and y-axis simply results in a fixed phase shift to 
the quadrature components. This suggests that the nature of fused pitch and roll 
in expressing a rotation is a property of the actual physical rotation, not 
whatever arbitrary reference frame is chosen to numerically quantify it. It can 
easily be demonstrated that Euler pitch and roll are not axisymmetric.

To demonstrate the fused pitch and roll axisymmetry mathematically, consider a 
robot undergoing the same rotation as in \figref{axisymmetry_rotation}. Once 
again, $\beta$ embodies the freedom of choice of the x and y-axis. We introduce 
the notation
\begin{equation}
\begin{aligned}
\rotb{U}{A}{R} &= \sub{T}{R}(\psi_0, \gamma_0, \alpha_0) = \sub{F}{R}(\psi_0, \theta_0, \phi_0, h_0), \\
\rotb{G}{B}{R} &= \sub{T}{R}(\psi_\beta, \gamma_\beta, \alpha_\beta) = \sub{F}{R}(\psi_\beta, \theta_\beta, \phi_\beta, h_\beta),
\end{aligned}
\eqnlabel{UAGGBRfusedtilt}
\end{equation}
for the tilt and fused angles representations of $\rotb{U}{A}{R}$ and 
$\rotb{G}{B}{R}$ respectively, where $\sub{T}{R}(\cdot)$ and $\sub{F}{R}(\cdot)$ 
is notation for the rotation matrices corresponding to the enclosed tilt and 
fused angles parameters, respectively. Using this notation, the previously 
established axisymmetry of fused yaw is equivalent to the statement
\begin{equation}
\psi_\beta = \psi_0. \eqnlabel{axisympsi}
\end{equation}
Substituting \eqnref{UGRpurez} into \eqnref{GBRexpansion}, and applying 
\eqnref{tiltfusedtorotmat} to $\rotb{U}{A}{R}$ gives
\begin{align}
\rotb{G}{B}{R} &= \rots{R}{z}(-\beta) \, \rotb{U}{A}{R} \, \rots{R}{z}(\beta) \notag\\
&=
\begin{bmatrix}
\cdot & \cdot & \cdot \\
\cdot & \cdot & \cdot \\
s_\beta s_{\phi_0} - c_\beta s_{\theta_0} & c_\beta s_{\phi_0} + s_\beta s_{\theta_0} & c_{\alpha_0}
\end{bmatrix}\mspace{-6mu},
\eqnlabel{GBRmatrix}
\end{align}
where the `$\cdot$' entries are omitted for brevity. Using 
\eqnref{tiltfusedtorotmat} to expand $\rotb{G}{B}{R}$, and comparing matrix 
entries to \eqnref{GBRmatrix}, gives
\begin{equation}
\mspace{-24mu}
\begin{aligned}
-s_{\theta_\beta} &= s_\beta s_{\phi_0} - c_\beta s_{\theta_0}, \\
s_{\phi_\beta} &= c_\beta s_{\phi_0} + s_\beta s_{\theta_0},
\end{aligned} \mspace{48mu}
c_{\alpha_\beta} = c_{\alpha_0}.
\eqnlabel{GBRcomparedcoeffs}
\end{equation}

\begin{figure}[!t]
\parbox{\linewidth}{\centering\includegraphics[width=1.0\linewidth]{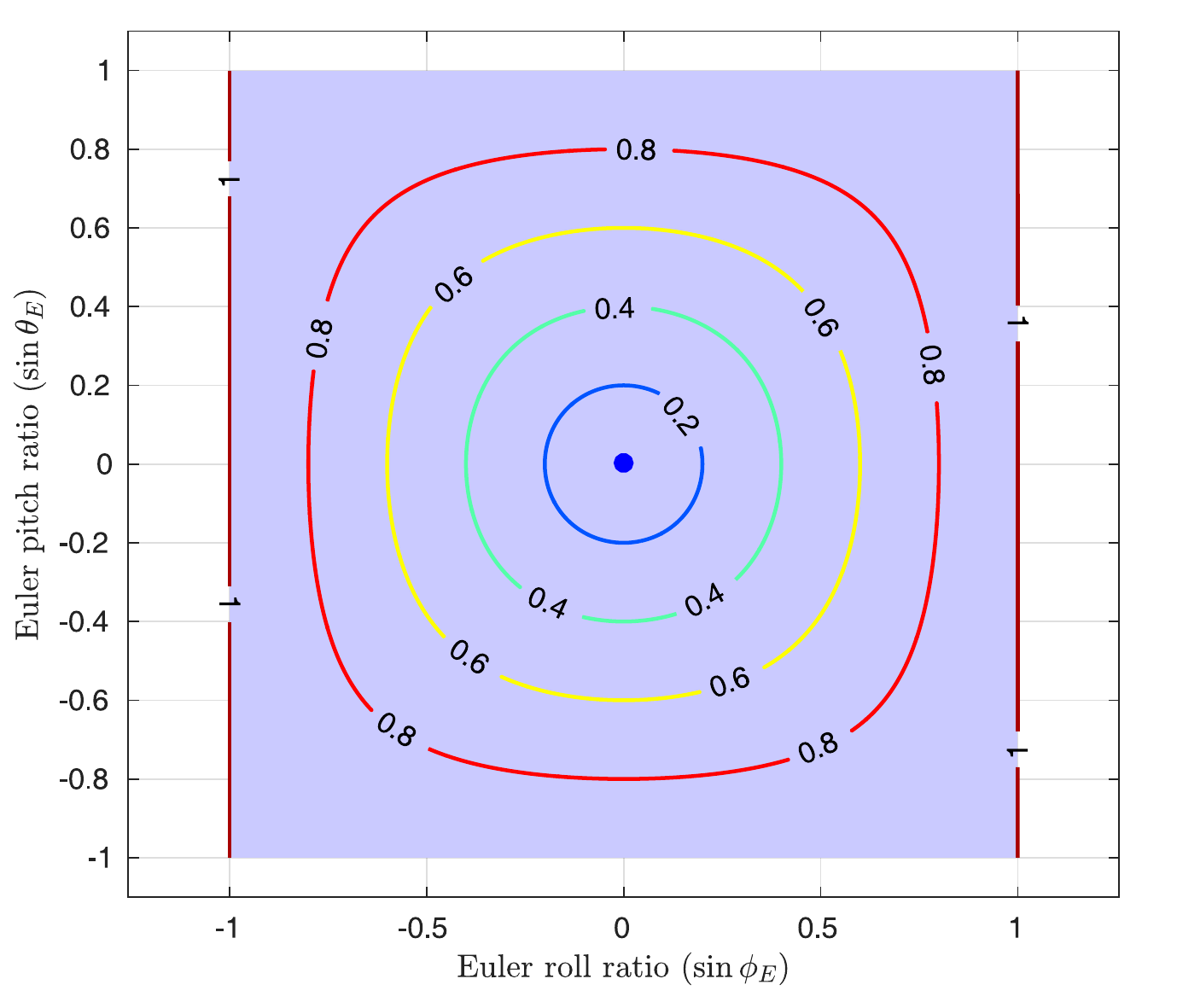}}
\caption{Level sets of constant $\sin\alpha$ in the Euler pitch ratio $\sin\thE$ 
vs.\ Euler roll ratio $\sin\phiE$ Cartesian space.}
\figlabel{euler_pitch_roll_axisymmetry}
\vspace{-2ex}
\end{figure}

\noindent This demonstrates the axisymmetry of the tilt angle $\alpha$,
\begin{equation}
\alpha_\beta = \alpha_0, \eqnlabel{axisymalpha}
\end{equation}
and as $h = \sign(c_\alpha)$, also the axisymmetry of $h$,
\begin{equation}
h_\beta = h_0. \eqnlabel{axisymhemi}
\end{equation}
\eqnref{GBRcomparedcoeffs} also leads to the matrix equation
\begin{equation}
\begin{bmatrix}
\sin\phi_\beta \\
\sin\theta_\beta
\end{bmatrix} =
\begin{bmatrix}
c_\beta & s_\beta \\
-s_\beta & c_\beta
\end{bmatrix}\mspace{-6mu}\begin{bmatrix}
\sin\phi_0 \\
\sin\theta_0
\end{bmatrix}\mspace{-6mu}.
\eqnlabel{sinbetaquadrature}
\end{equation}
By identifying the middle matrix as a 2D rotation matrix that rotates clockwise 
by $\beta$, this equation can be seen to be the mathematical expression that 
epitomises the axisymmetry of fused pitch and roll, in the sense that they vary 
in a rotational manner proportional to the choice of $\beta$. The effect of 
varying $\beta$, and how this leads to a uniform circular locus of sine ratios 
$(\sin\phi_\beta, \sin\theta_\beta)$, is illustrated in 
\figref{pitch_roll_beta}. From \eqnref{sinbetaquadrature}, the phase shift to 
the quadrature sinusoid components can be seen to be $-\beta$. In consideration 
of \eqnref{fusedpitchrollquadrature}, this yields the relation
\begin{equation}
\gamma_\beta = \gamma_0 - \beta. \eqnlabel{axisymgamma}
\end{equation}
This is an expression of the axisymmetry of the tilt axis angle $\gamma$, 
equivalent to that for fused pitch and roll. As $(\gamma,\alpha)$ completely 
parameterises the tilt rotation space $(\theta,\phi,h)$, it can be seen from 
\eqnref{axisymalpha} and \eqnref{axisymgamma} that all possible loci of sine 
ratios $(\sin\phi_\beta, \sin\theta_\beta)$ as $\beta$ varies can be plotted by 
examining the contours of constant $\alpha$ while $\gamma$ varies. This is 
equivalent to generating the level sets of constant $\sin\alpha$ in the fused 
pitch ratio vs.\ fused roll ratio plane, the result of which is shown in 
\figref{pitch_roll_axisymmetry}. The axisymmetry of fused pitch and roll can be 
clearly visually identified in the figure. An analogous plot for Euler angles 
is provided in \figref{euler_pitch_roll_axisymmetry}. The non-axisymmetry of the 
Euler pitch and roll is clearly visible.

\begin{figure}[!t]
\parbox{\linewidth}{\centering\includegraphics[width=1.0\linewidth]{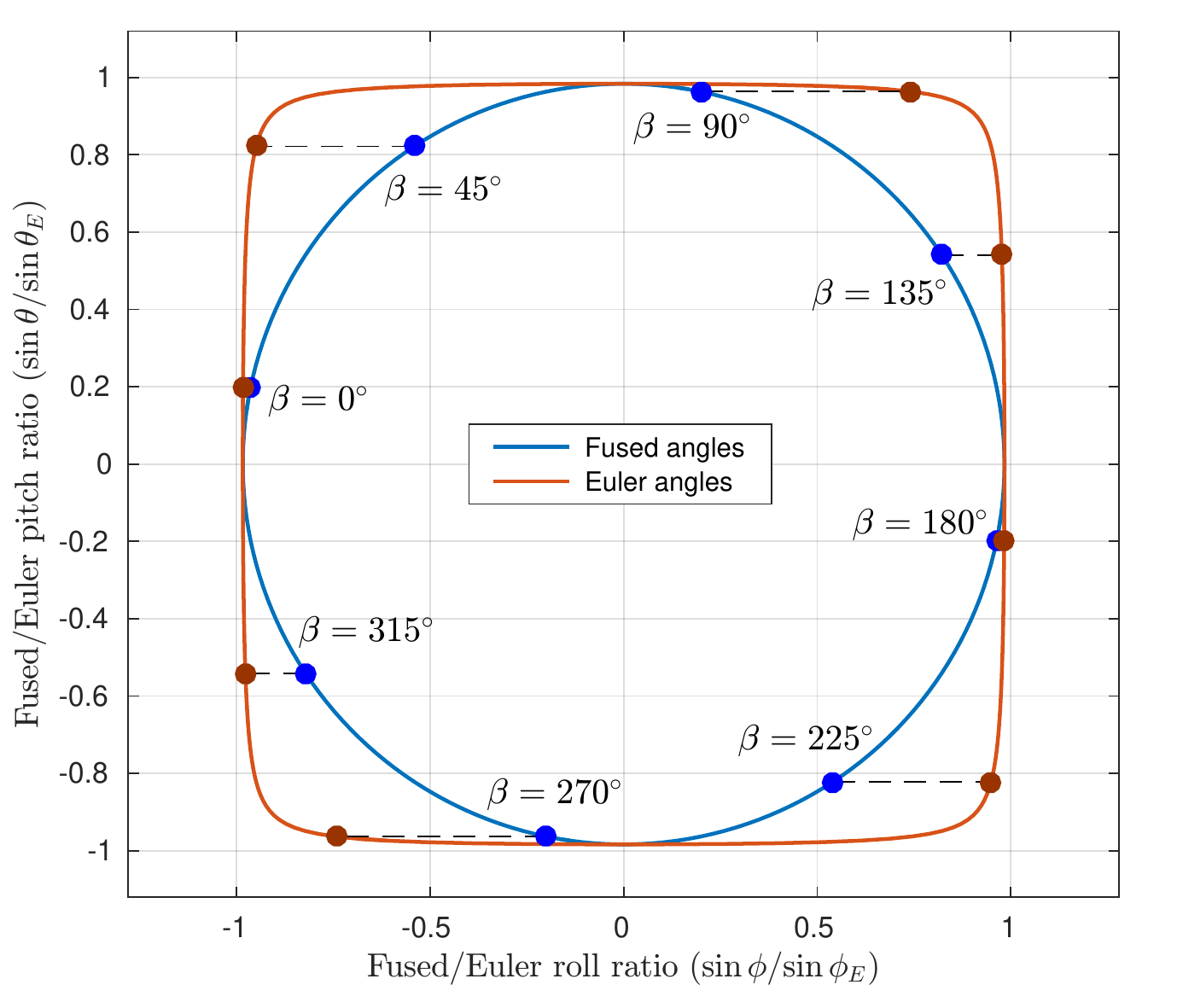}}
\caption{Loci of the sine ratios $(\sin\phi_\beta, \sin\theta_\beta)$ as $\beta$ 
varies, i.e.\ for all possible choices of x and y-axis, once (inner curve) for 
fused pitch $\theta$ and roll $\phi$, and once (outer curve) for Euler pitch 
$\thE$ and roll $\phiE$. The base rotation $\rotb{U}{A}{R}$ used for both loci 
is $\sub{F}{R}(-1.2, 0.2, -1.3, -1)$. The non-circularity of the Euler locus, as 
well as the non-uniformity of the associated keypoints, demonstrate the 
violation of axisymmetry for Euler pitch and roll.}
\figlabel{euler_pitch_roll_beta}
\vspace{-2ex}
\end{figure}

The non-axisymmetry of all three Euler angles parameters is further visualised 
in \figref{euler_pitch_roll_beta} and \figref{euler_param_beta}. The two figures 
also illustrate the corresponding axisymmetry of the fused angles parameters for 
the same base rotation $\rotb{U}{A}{R}$. Conceptually, the problem of Euler 
pitch and roll is the fundamental requirement of a defined order of rotations. 
As can be identified in \figref{sensitivity}, this leads to definitions of pitch 
and roll that do not correspond to each other in behaviour, as one then 
implicitly depends on the value of the other.

\section{Conclusion}
\seclabel{conclusion}

As has been shown in detail, fused angles possess many important properties that 
Euler angles do not. These properties relate to the nature of the singularities, 
the axisymmetry of the parameters, and the absence of any parameter 
interdependencies. As a result, fused angles not only intuitively quantify the 
amount of rotation within the three major planes, as was the core objective,
but also fulfil natural and intuitive expectations about how a rotation 
representation should behave, especially for the application of representing the 
orientation of a balancing body \cite{Allgeuer2016a}.

\begin{figure}[!t]
\parbox{\linewidth}{\centering\includegraphics[width=1.0\linewidth]{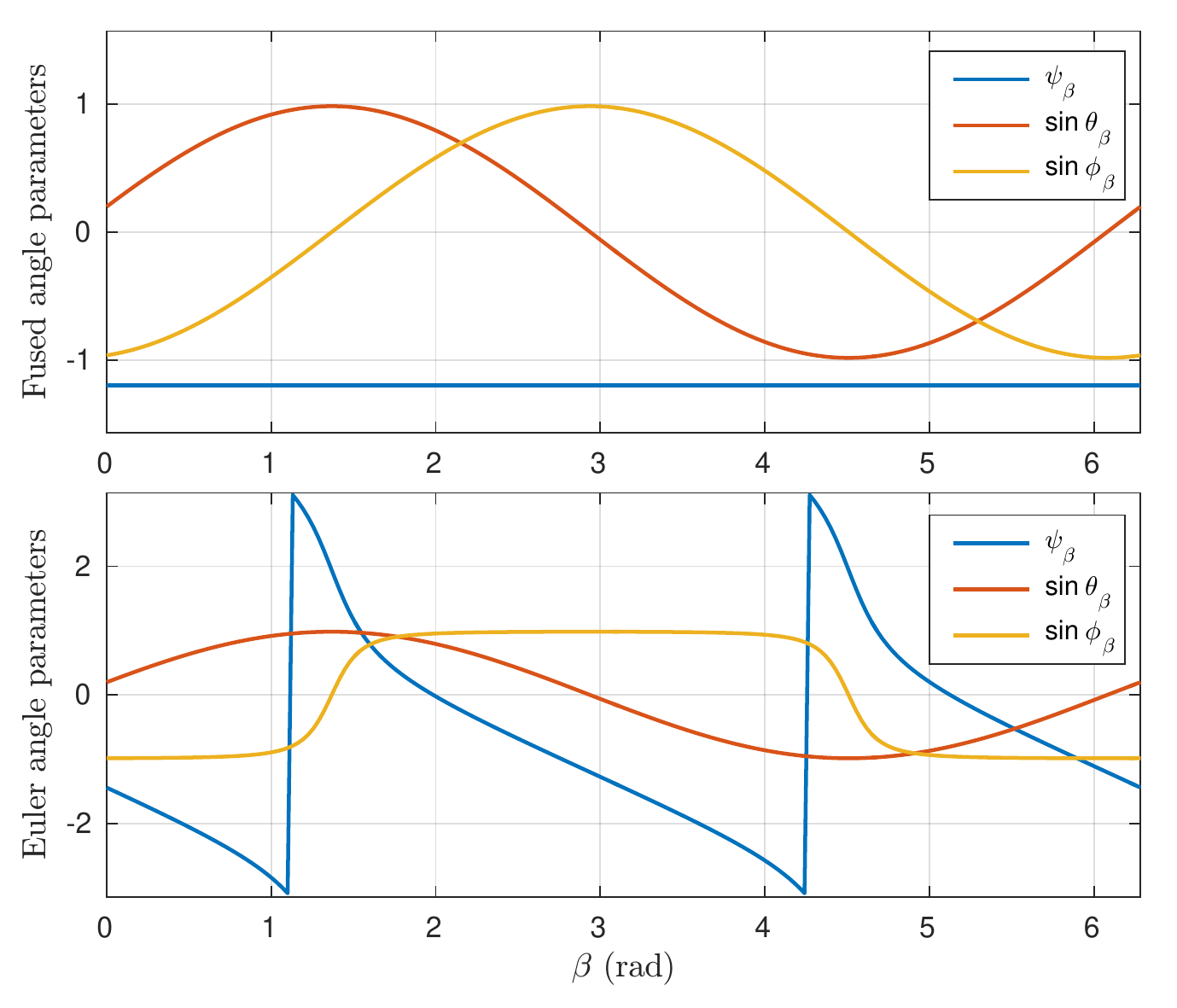}}
\caption{Plots of yaw, and the pitch and roll sine ratios, against $\beta$ for both 
fused angles (top) and Euler angles (bottom). The base rotation $\rotb{U}{A}{R}$ 
used for both plots is $\sub{F}{R}(-1.2, 0.2, -1.3, -1)$, as in 
\figref{euler_pitch_roll_beta}. The invariance of the fused yaw, as well as the 
exact quadrature nature of the fused pitch and roll can be clearly identified. 
The non-axisymmetry of Euler pitch and roll, and in particular the irregularity 
of Euler yaw, can also be seen.}
\figlabel{euler_param_beta}
\vspace{-2ex}
\end{figure}
\bibliographystyle{IEEEtran}
\bibliography{IEEEabrv,ms}

\end{allowdisplaybreaks}
\end{document}